\documentclass[preprint,12pt]{elsarticle}

\usepackage{lineno}
\usepackage{graphicx}
\usepackage{pgfplots}
\usepackage{pgfplotstable}
\pgfplotsset{compat=1.17}
\usepackage{amsmath,amssymb,amsfonts}
\usepackage{algorithmic}
\usepackage{graphicx}
\usepackage{tabularx,subcaption,caption,adjustbox,multicol,multirow,booktabs}
\usepackage{colortbl}
\usepackage{xcolor}
\usepackage{svg}
\usepackage{textcomp}
\usepackage{hyperref}
\usepackage{rotating}
\usepackage{array}
\journal{Computers in Biology and Medicine}
\begin{document}

\begin{frontmatter}

\title{Sam2Rad: A Segmentation Model for Medical Images with Learnable Prompts}
\affiliation[uofa]{organization={Department of Radiology and Diagnostic Imaging, University of Alberta},
            country={Canada}}

\affiliation[mcgill]{organization={School of Computer Science, McGill University},
          country={Canada}}

\author[uofa]{Assefa Seyoum Wahd}
\author[uofa]{Banafshe Felfeliyan}
\author[uofa]{Yuyue Zhou}
\author[uofa]{Shrimanti Ghosh}
\author[uofa]{Adam McArthur}
\author[mcgill]{Jiechen Zhang}
\author[uofa]{Jacob L. Jaremko}
\author[uofa]{Abhilash Hareendranathan}

\begin{abstract}
    The use of foundation models like the Segment Anything Model (SAM) for medical image segmentation requires high-quality manual prompts, which are time-consuming to generate and require medical expertise. Even when prompted with sparse prompts, like boxes, points, or text, and dense prompts such as masks SAM or its variants like MedSAM fine-tuned on medical images fail to segment bones in ultrasound (US) images due to significant domain shift. 

    We propose a new prompt learning approach to adapt SAM, its recent variant SAM 2, and other variants to segment bony regions in US images without requiring human prompts. We introduce a prompt predictor network (PPN) with a lightweight cross-attention module to augment the existing prompt encoder, predicting prompt embeddings directly from features extracted by the image encoder. PPN outputs bounding box and mask prompts and $N$ 256-dimensional embeddings for the regions of interest. Our new framework also allows optional manual prompting which can be concatenated with learned prompts as inputs to the the mask decoder. PPN and the mask decoder can be trained end-to-end using parameter-efficient fine-tuning (PEFT) methods. To preserve SAM's vast world knowledge, we ideally want to keep all modules in SAM frozen and train only the prompt predictor network. We demonstrate the prompt predictor network's effectiveness by freezing all parameters in SAM and achieving a Dice score comparable to that of the fine-tuned mask decoder.  

     Our model can be used autonomously without human supervision, semi-autonomously with \textit{human-in-the-loop}, or fully manual where the model uses human prompts only, like the original SAM. In the semi-autonomous mode, the model predicts masks, and if unsatisfactory, users can provide additional prompts (boxes, points, or brushes) to refine the prediction.  The autonomous setting can be used for real-time applications while the semi-autonomous is ideal for data labeling, or in active learning frameworks. 

     We tested the proposed model - Sam2Rad on 3 musculoskeletal US datasets - wrist (3822 images), rotator cuff (1605 images), and hip (4849 images). 

     Without Sam2Rad, all SAM2 variants failed to segment shoulder US in zero-shot generalization with bounding box prompts, while SAM2 showed better segmentation accuracy across other US datasets vs. SAM.  Our model, Sam2Rad, improved the performance of all SAM base networks in all datasets, without requiring manual prompts.  The improvement in accuracy ranged from a 2-7\% increase in Dice score for hip or wrist, and up to 33\% improvement in Dice score (from 49\% to 82\%) on shoulder data.  Notably, Sam2Rad could be trained with as few as 10 labeled images. Sam2Rad is compatible with any SAM architecture and can be utilized for automatic segmentation.  
     
    The code is available at \url{https://github.com/aswahd/SamRadiology}.

   \end{abstract}


\end{frontmatter}

\section{Introduction}
\label{introduction}

Automated segmentation of structures in medical images is an essential step for various computer-aided diagnoses and treatment planning procedures. Over the past decade, the evolution of deep learning models has significantly accelerated the progress of automatic medical image segmentation. Traditionally, segmentation tasks are approached separately by specialized networks \citep{ronnebergerUNetConvolutionalNetworks2015a, heMaskRCNN2018, isenseeNnUNetSelfadaptingFramework2018, zhouUNetNestedUNet2018, oktayAttentionUNetLearning2018,  chenTransUNetTransformersMake2021,  hatamizadehUNETRTransformers3D2021,  caoSwinUnetUnetlikePure2021}. As a result, their use in broader clinical contexts is limited as this approach requires substantial amounts of data to train each task and ignores correlations between different segmentation objectives. Meanwhile, deep learning scaling laws suggest that increasing the size of the training dataset and the number of parameters of a model improves accuracy \citep{kaplanScalingLawsNeural2020, rosenfeldConstructivePredictionGeneralization2019,bahriExplainingNeuralScaling2024a, hernandezScalingLawsTransfer2021,Zhai_2022_CVPR} | resulting in foundation models. These models, often trained on extensive datasets with a wide array of general information, encapsulate a broad spectrum of world knowledge that can be adapted to specific applications with relatively minimal effort compared to training a model from scratch. Foundation models are shown to generalize well to downstream tasks with few labeled images, are more robust to adversarial attacks, resilient to domain shift, and effectively work in zero-shot settings. Consequently, foundation models have become popular in medical imaging, where large-scale labeled datasets are often scarce.

\begin{figure}[!t]
     \centering
     \includegraphics[width=\linewidth]{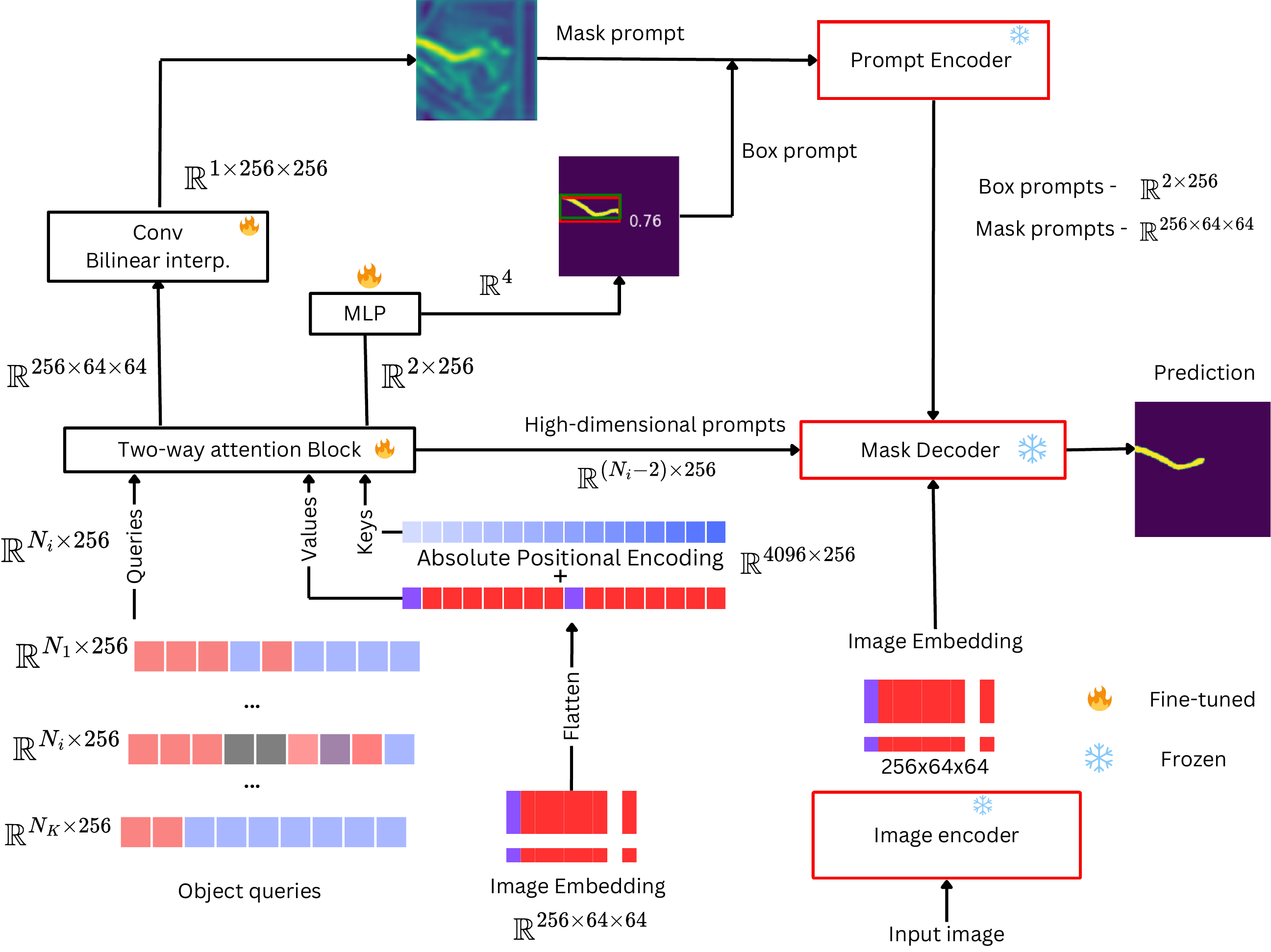}
     \caption{\textbf{Sam2Rad architecture}. We propose a lightweight two-way attention module to predict prompts for a queried object. Each object ($k$) is represented by $T_k \in \mathbb{R}^{N_i\times 256}$ learnable prompts. These class prompts are conditioned on features obtained from a pretrained SAM image encoder. During inference, the prompt encoder can be removed to run the model autonomously, or it can be used in combination with the prompt predictor network for human-in-the-loop operation.}
     \label{fig:model_architecture}
 \end{figure}

\begin{figure}[!tbh] 
    \centering
    \includegraphics[height=0.72\textheight]{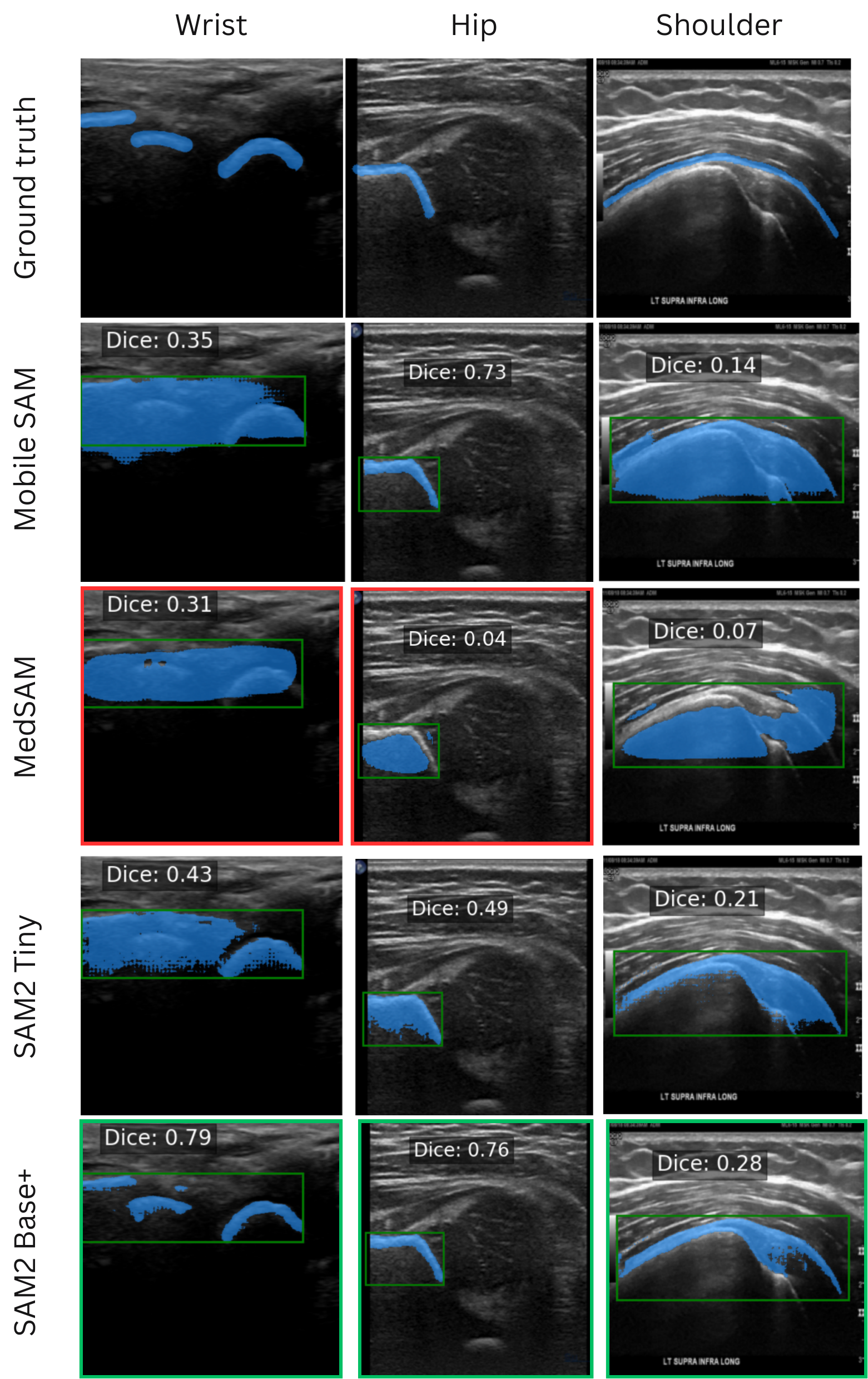}
    \caption{ 
        SAM variants show clear limitations when segmenting US images using bounding box prompts. Surprisingly, MedSAM \citep{maSegmentAnythingMedical2023}, despite being trained on a large-scale medical dataset, performs the worst. In contrast, SAM2 \citep{raviSAMSegmentAnything}, SAM's latest iteration, demonstrates the best zero-shot generalization. However, smaller SAM2 variants (such as Sam2-Tiny) still incorrectly include background in the segmentation mask. Notably, all models struggle with shoulder US. These observations suggest that bounding box prompts may not be sufficient, or that the models require fine-tuning. In section \ref{prompt_learning}, we demonstrate that our proposed prompt learning approach can improve the performance of these models while keeping their pretrained parameters frozen.
    }
   \label{fig:sam_fails}
\end{figure}

\section{Related Works}
\subsection{Foundation Models}
 Foundation models, popularized by OpenAI's CLIP \citep{wangMedCLIPContrastiveLearning2022}, are large-scale deep learning models trained on extensive datasets, enabling them to effectively generalize across various downstream tasks. These models are typically pretrained using self-supervised learning methods on large datasets, allowing them to learn a wide range of patterns without requiring labeled examples. CLIP, for instance, is trained on millions of text-image pairs to create a joint embedding space for images and text. It has been widely used for image classification with text prompts, object detection, image segmentation, and various NLP tasks. Segment Anything Model (SAM) is a foundation model trained on the largest segmentation dataset to date, comprising over 10 million images and 1 billion masks.
 
Segment Anything Model (SAM) has emerged as a foundation model for natural image segmentation, offering one framework for handling multiple segmentation tasks at once, given any segmentation prompt, such as points, bounding boxes, masks or text ~\citep{kirillovSegmentAnything2023}. 
SAM has inspired a plethora of follow-up works, leading to the development of universal models for medical image segmentation ~\citep{maSegmentAnythingMedical2023,zhangSegmentAnythingModel2023,dengSegmentAnythingModel2023,mazurowskiSegmentAnythingModel2023,huBreastSAMStudySegment2023,ConvolutionMeetsLoRA2023,customized_zhang_2023,segment_zhang_2023,guHowBuildBest2024,leiMedLSAMLocalizeSegment2023,wuMedicalSAMAdapter2023,yeSAMed2D20MDatasetSegment,sam_chen_2023,chenSAMAdapterAdaptingSegment2023a,putzSegmentAnythingFoundation2023,ningPotentialSegmentAnything2023,linSAMUSAdaptingSegment2023}. SAM2 \cite{raviSAMSegmentAnything}, the latest iteration of SAM, is designed for both image and video segmentation. It can track objects across video frames when given a prompt in any frame. Notably, SAM2 operates $6\times$ faster than its predecessor. It incorporates a memory attention mechanism, allowing the current frame's prediction to be conditioned on the previous frames' image embeddings and predictions.

However, studies that explored SAM's application in the medical image domain have identified its limitations in specific unseen medical modalities ~\citep{mazurowskiSegmentAnythingModel2023,chenSAMFailsSegment2023,shiGeneralistVisionFoundation2023,roySAMMDZeroshot2023,huangSegmentAnythingModel2024a}. Efforts have been made to address this by fine-tuning SAM on large curated datasets in medical domain, resulting in medical foundation models like ~\citep{maSegmentAnythingMedical2023,chengSAMMed2D2023}. Nonetheless, these models' performance remains limited to the modalities they were trained on and heavily relies on prompt quality. This poses a significant challenge when these models are applied to substantially different data distributions such as ultrasound (see Figure \ref{fig:sam_fails}). As shown in the Figure, SAM and its variants have sub-optimal performance on US data. This could be due to the unique nature of imaging artifacts seen in US, like the speckle interference pattern, which occurs when the distance between adjacent scatters is less than the cell resolution. Addressing such limitations would require domain-specific fine-tuning for optimal performance while maintaining broad applicability. Furthermore, these foundation models require manual prompts, which are practically difficult without the necessary medical expertise. As a result, deep learning methods for automatically generating appropriate prompts have become a significant focus of research.

\subsection{Prompt Learning} 

Traditional approaches to training deep learning models often involve either training the model from scratch or pretraining it with unlabeled data followed by fine-tuning with labeled data for specific tasks. In contrast, prompt-based learning models can autonomously adapt to various tasks by utilizing cues provided by users. These models leverage the domain knowledge introduced through prompts to guide their performance, enabling more efficient and flexible task adaptation without extensive task-specific training.

SAM demonstrates impressive zero-shot generalization to natural images. However, it requires prompts—such as points, boxes, text, or masks—to segment objects in an image, even during inference. Prompts can be either provided by humans (e.g., drawing boxes around objects or inputting text) or predicted by standalone models. Grounded SAM \citep{renGroundedSAMAssembling2024} uses Grounding DINO \citep{liuGroundingDINOMarrying2023} to generate the bounding boxes for all objects in a given image,  which are then used to prompt SAM. The limitation of this approach is that it requires a model with a separate image encoder trained from scratch. 
 
Recently, automatic techniques like PerSAM \citep{zhangPersonalizeSegmentAnything2023} and Matcher \citep{liuMatcherSegmentAnything2023}  have been proposed to generate bounding box prompts for natural images. These approaches assume that foreground and background patches are dissimilar and use patch similarities between a reference image and a test image to generate prompts. However, in US images foreground and background patches are often hard to distinguish, resulting in a high cosine similarity between unrelated patches. We propose a new method to learn prompts using SAM's image encoder instead of relying on other standalone models like Grounding DINO \citep{liuGroundingDINOMarrying2023}. We develop a prompt predictor network (PPN) that takes in image features extracted by SAM's image encoder and predicts appropriate prompts for regions of interest. 

Instead, we propose a lightweight attention mechanism to learn a non-linear function to predict the location of the target regions in the image. The prompt predictor network uses a lightweight attention module that operates on the image features obtained from SAM's image encoder to predict the prompts for the target regions. 

To the best of our knowledge, this is the first work to report prompt learning for SAM2 \citep{raviSAMSegmentAnything}, the latest version of SAM. For a comparison between our prompt predictor network and other prompting techniques, see Figure \ref{fig:prompt_learning_comparison}.

\subsection{Parameter-efficient Fine-tuning}
Several studies have shown that SAM fails to generalize to medical images even with manual prompts \citep{mazurowskiSegmentAnythingModel2023,chenSAMFailsSegment2023,shiGeneralistVisionFoundation2023,roySAMMDZeroshot2023,huangSegmentAnythingModel2024a}. We encountered similar issues with ultrasound (US) images, as illustrated in Figure \ref{fig:sam_fails}.  While a straightforward approach to address this issue is fine-tuning all the model's parameters|vanilla fine-tuning, this method can be suboptimal, often requiring a substantial amount of labeled data and sometimes performing poorly compared to parameter-efficient fine-tuning (PEFT) methods. For example, CoOp \citep{zhouConditionalPromptLearning2022} reported a 40\% drop in classification accuracy when vanilla fine-tuning the CLIP image encoder. PEFT is a more efficient way to adapt foundation models to downstream tasks. PEFT methods like adapters \citep{wuMedicalSAMAdapter2023} and low-rank adaptation (LoRA) \citep{huLoRALowRankAdaptation2021} are designed to adapt foundation models to downstream tasks with minimal changes to the model's parameters. PEFT methods are particularly useful when labeled data is limited. Adapters \citep{wuMedicalSAMAdapter2023} insert small modules in parallel to the transformer layers of a foundation model and fine-tune only these new parameters. LoRA \citep{customized_zhang_2023} is another PEFT method that fine-tunes only a small subset of the model's parameters. To avoid SAM or SAM2 failure in image segmentation even with manual prompts, the mask decoder can be fine-tuned using LoRA while keeping the image encoder's parameters frozen. Studies show integrating PEFT with SAM not only reduces computational costs but also mitigates overfitting and improves the overall segmentation accuracy \citep{guHowBuildBest2024}.

\subsection{Key Contributions}
We propose a new PPN that directly predicts $N$ 256-dimensional embeddings alongside bounding box coordinates and mask prompts. This approach allows the network to learn an abstract representation of a prompt instead of a sparse representation like a box. Moreover, $N$ can be set to any value, in contrast to other methods that typically predict a single bounding box per object. Studies show that adapting SAM to medical imaging often requires several-point prompts. In PPN, $N$ can represent any number of points or other object-relevant information. For instance, for objects that are challenging to segment, such as thin overlapping tissues, the network can adapt by predicting multiple point prompts or using alternative representations

To the best of our knowledge, this is the first work to report prompt learning for SAM2 \citep{raviSAMSegmentAnything}, the latest version of SAM. 

The code is available on \url{https://github.com/aswahd/Sam2Radiology}.

\section{Methodology}
\label{methodology}

We propose a novel framework, SAM2Rad, designed for segmenting medical images without the need for manual prompts. This paper focuses specifically on applying this framework to ultrasound imaging. SAM2Rad incorporates a prompt predictor (generator) network (PPN) that predicts candidate bounding box proposals, mask prompts, and $N$ 256-dimensional prompts. Similar to object detection in Mask R-CNN \citep{heMaskRCNN2018}, where bounding box regression heads use features from the backbone network to predict coordinates, PPN's bounding box regression head can be viewed as object detection. It utilizes features from SAM's image encoder to generate appropriate prompts for segmenting target regions, functioning analogously to Mask R-CNN's region proposal network (RPN) \citep{heMaskRCNN2018}.

\subsection{Sam2Rad}
The proposed SAM2Rad framework comprises a pretrained SAM/SAM2 encoder, mask decoder, prompt encoder, and prompt predictor network (Figure \ref{fig:model_architecture}).

\paragraph{Image encoder} The image encoder extracts features from the input image. It is based on the vision transformer (ViT) architecture \citep{dosovitskiyImageWorth16x162020} and pretrained using masked autoencoders (MAE) \citep{heMaskedAutoencodersAre2021}. For an input image of size $3\times 1024 \times 1024$, the image encoder outputs embeddings of size $256 \times 64 \times 64$.

\paragraph{Mask decoder} The mask decoder transforms the image embeddings and prompt embeddings to a final mask.

\paragraph{Prompt encoder} The prompt encoder processes both sparse prompts (points and bounding boxes) and dense prompts (masks). Points are encoded into 256-dimensional embeddings, combining positional and learned embeddings. Bounding boxes are similarly encoded, using positional encodings for the top-left corner and learned embeddings for the bottom-right corner. Masks ($256\times 256$) are transformed into $256 \times 64 \times 64$ embeddings using a $4\times$ downsampling convolution block.

\begin{figure}[!tbh]
     \centering
     \includegraphics*[width=\textwidth]{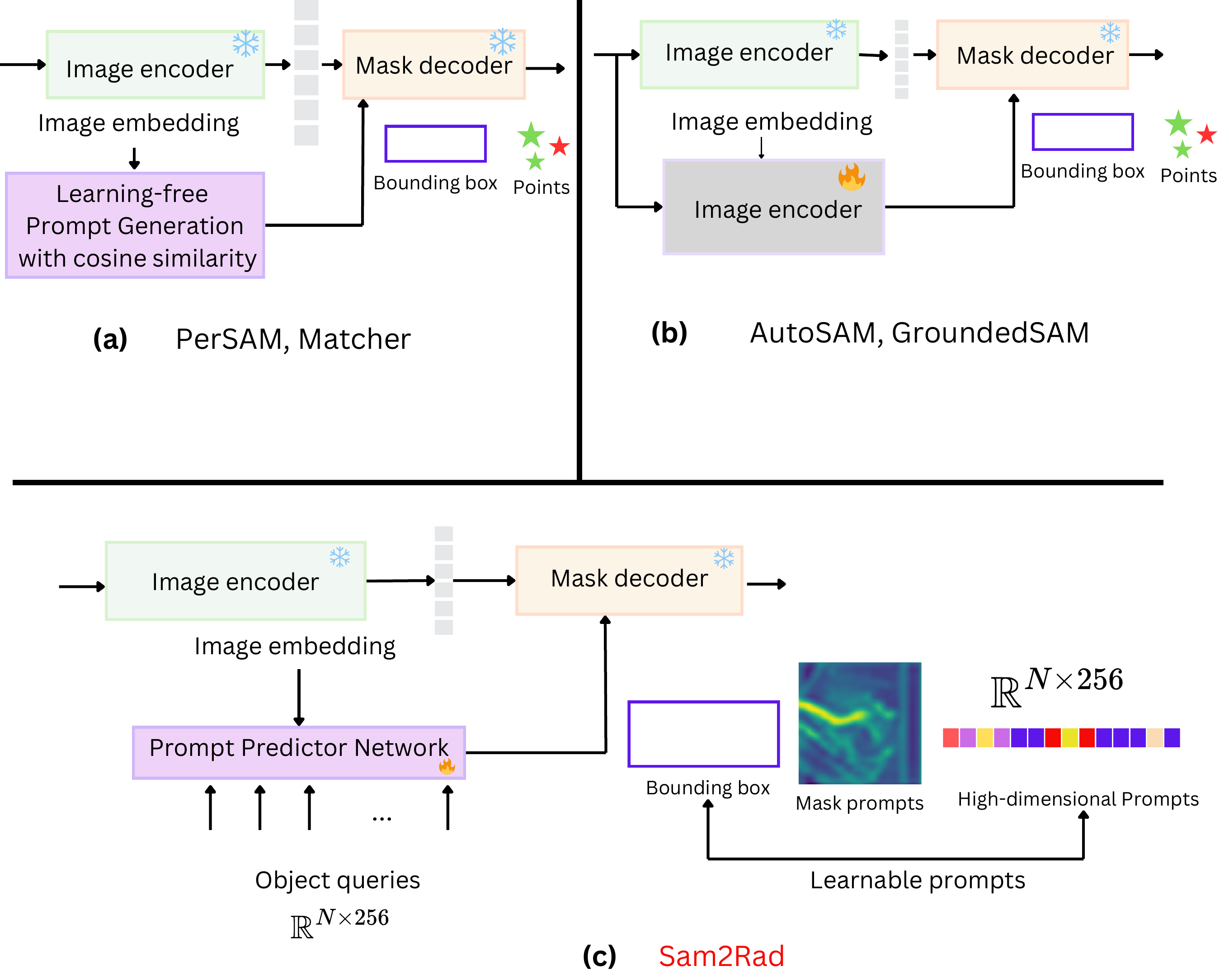}
     \caption{\textbf{Comparison between our prompt predictor network and others.} PerSAM \citep{zhangPersonalizeSegmentAnything2023} and Matcher \citep{liuMatcherSegmentAnything2023} use learning-free cosine similarity to predict box and point coordinates by comparing image patches between a reference image and a test image \textbf{(a)}. Given image features, our \textit{prompt predictor network} predicts prompts for the queried object. The predictor network outputs bounding box coordinates of target regions, an intermediate mask prompt, and $N$ high-dimensional learnable embeddings ($\mathbb{R}^{N\times 256}$). All the predicted prompts (learnable prompts) are then fed to SAM's mask decoder to generate a mask. The prompt predictor network can handle multiple classes by providing learnable class queries \textbf{(c)}. In contrast, \citep{liSelfPromptingLargeLanguage2023, xieMaskSAMAutopromptSAM2024} predict only bounding boxes or intermediate masks, while GroundedSAM \citep{renGroundedSAMAssembling2024} and AutoSAM \citep{shaharabanyAutoSAMAdaptingSAM2023a} predict prompts using a standalone image encoder \textbf{(b)}.}
     \label{fig:prompt_learning_comparison}
\end{figure}

To align the learned prompts with the manual prompts SAM was trained on, our predictor network outputs bounding box coordinates, mask prompts, and $N$ 256-dimensional embeddings. For a comparison between our prompt predictor network and other prompting techniques, see Figure \ref{fig:prompt_learning_comparison}.

In the next section, we outline the design of the prompt predictor.

\subsubsection{Prompt predictor network (prompt learning)}
\label{prompt_learning}

\begin{figure}[!t]
\centering
\includegraphics[width=\textwidth]{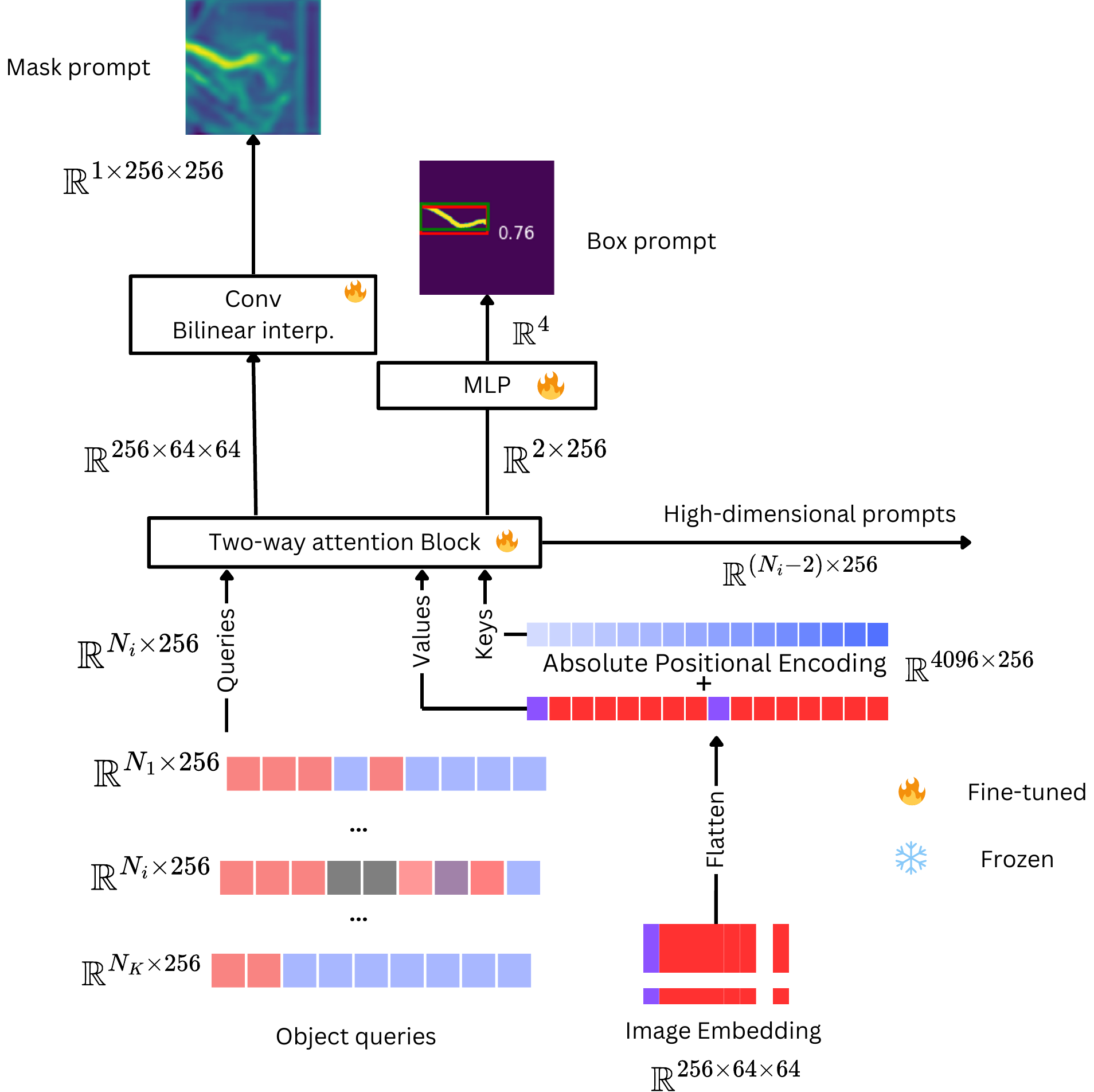}
\caption{\textbf{Prompt predictor network architecture.} Given image features, our \textit{prompt predictor network} predicts prompts for the queried object. The predictor network outputs bounding box coordinates of the target object, an intermediate mask prompt, and $N$ 256-dimensional learnable embeddings ($\mathbb{R}^{N\times 256}$). All the predicted (learnable) prompts are then fed to SAM's mask decoder to generate a mask. The prompt predictor network can handle multiple classes by providing distinct class queries.} 
\label{fig:ppn_architecture}
\end{figure}

 Given an image $I$ of size $\mathbb{R}^{3\times H \times W}$, the SAM image encoder provides embeddings $E \in \mathbb{R}^{C\times H' \times W'}$.
 Each point in $E$ corresponds to a patch in $I$. PerSAM \citep{zhangPersonalizeSegmentAnything2023} and Matcher \citep{liuMatcherSegmentAnything2023} use this feature map to compute the cosine similarity between $E$ and a reference image's embeddings $E_r \in \mathbb{R}^{C\times H' \times W'}$. 
 \par As mentioned in the introduction\textsuperscript{\ref{introduction}}, US images often exhibit significant similarities between foreground and background patches, resulting in a high cosine similarity between unrelated patches. So, instead of using the cosine similarity, we propose a lightweight attention mechanism to learn a non-linear function to predict the location of the target regions in $I$, given the patch embeddings of $E$. To this end, we represent the learnable positional embeddings (tokens) as $T \in \mathbb{R}^{N\times C}$, where $N$ is the number of learnable tokens, and $C$ is the number of channels. 
 \par Since the target regions may appear at arbitrary locations in the image $I$, the learnable position embeddings ($T$) need to be dynamically updated to incorporate information about the object's location. First, we add positional encodings to the patch embeddings in $E$ as shown in Figure \ref{fig:ppn_architecture}. The updated image embeddings and learnable tokens are then passed to a cross-attention module, where the learnable tokens are the queries and keys/values come from the image embeddings, resulting in updated tokens $\hat{T} \in \mathbb{R}^{N\times C}$. 
 \par The image embeddings are also updated by attending to the updated learnable tokens, resulting in image tokens of shape $\hat{E} \in \mathbb{R}^{4096\times 256}$. To align the learned prompts with the prompts that SAM was trained on (box, mask, and point), $\hat{T}$ is split into two parts: the first two tokens ($\mathbb{R}^{2 \times 256}$) are used to predict bounding box coordinates, and the remaining $(N-2)$ tokens are used as high dimensional prompts. The updated image tokens are used to generate a mask prompt of shape $\mathbb{R}^{1 \times 256 \times 256}$. The predicted bounding box, mask and  $(N-2)$ 256-dimensional prompts are then fed to SAM's mask decoder to generate the final mask. The prompt predictor network can handle multiple classes by providing distinct class queries. $\hat T \in \mathbb{R}^{(N-2) \times 256}$ may represent any combination of prompts suitable for the task at hand; for example, some shapes are easier to segment with boxes while others are easier with several-point prompts. The high-dimensional prompts will decide whether to use point/box prompts or semantic information based on the properties of the dataset.

\subsection{Training Losses} 
In some cases, the queried object may not be present in the image, such as in background images or when a specific class is missing in multi-class segmentation. To address this, we introduce a fully connected object prediction head and an output token within the mask decoder. During inference, if the object prediction head outputs a value less than 0, we disregard the segmentation output.

Given an image $I \in \mathbb{R}^{3\times H\times W}$, a binary ground-truth mask $y \in \mathbb{R}^{H \times W}$, we obtain bounding box coordinates $\mathbb{R}^4$, mask prompts $\mathbb{R}^{1\times 256 \times 256}$, and a predicted mask $\hat y \in \mathbb{R}^{H \times W}$ (after sigmoid activation), the training loss function is formulated as follows:

\begin{equation}
     \mathcal{L}_{\text{Total}} = \lambda_1 \mathcal{L}_{\text{Mask}} + \lambda_2 \mathcal{L}_{\text{Box}} + \lambda_3 \mathcal{L}_{\text{Objectness}},
     \label{eqn:loss_fn}
\end{equation}
where 
$\mathcal{L}_\text{mask}$ is the sum of focal loss \citep{linFocalLossDense2018} for the mask prompt, and the final mask prediction, where the focal loss is 
\begin{equation}
     \mathcal{L}_{\text{Focal}}(\hat y_i, y_i) =  -y_i(1 - \hat y_i)^\gamma\log(\hat y_i) - 
     \alpha (1 - y_i) {\hat y_i}^\gamma \log(1 - \hat y_i),
     \label{eqn:focal_loss}
\end{equation}
$\mathcal{L}_\text{Box}$ is a linear combination of L1 loss and the generalized IoU loss \citep{rezatofighiGeneralizedIntersectionUnion2019} for bounding box regression, and $\mathcal{L}_\text{Objectness}$ is the binary cross-entropy loss to predict the presence of the queried object in the image.

\section{Experiments}
\label{experiments}

 \subsection{Datasets} The datasets were curated from the US scans of different body parts including wrist, elbow, hip, and shoulder (Table \ref{tab:num_train_test}) obtained with institutional ethics board approval. Data was presented to the networks in anonymized form.  After splitting the subjects into training/test sets based on subject study ID, we extracted 2D frames from each volume and filtered out frames with no foreground objects.
\begin{table}[!t]
     \centering
     \caption{Number of images in the train/test sets.}
     \begin{tabular}{c c c c}
         \toprule
          Dataset & \# of training images & \# of test images & \# of subjects\\ 
          \toprule
          Hip & 4987 & 4849 & 50\\
          Wrist & 3738 & 3822 & 101 \\
          Shoulder (Bursa) & 7364 & 1605 & 206\\
          \toprule
     \end{tabular}
     \label{tab:num_train_test}
 \end{table}

 \subsection{Data Augmentation}
 \label{sec:data_augmentation}
\begin{figure}[!t]
     \centering
     \includegraphics[width=\textwidth]{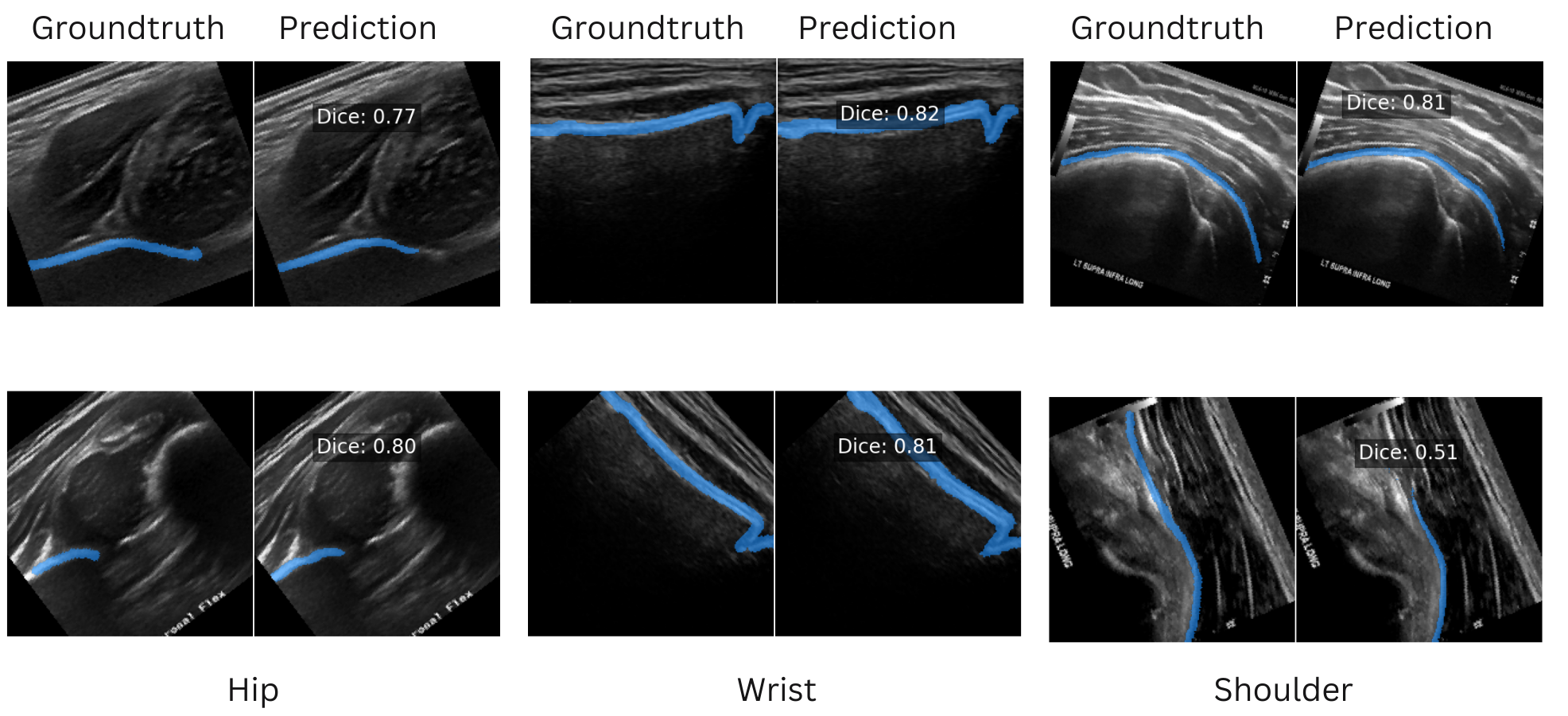}
     \caption{The model is robust to strong augmentations. For details on the data augmentation techniques applied during training, see section \ref{sec:data_augmentation}.}
     \label{fig:data_augmentation}
\end{figure}

 Overparametrized networks come with the risk of overfitting, especially when working with limited training data.  To enhance the model's generalization and prevent overfitting, we apply strong augmentations: random horizontal/vertical flip, random 20\% horizontal/vertical translations, a random rotation in the range [-90, 90] degrees, and a random resized crop in the range of $[0.8, 1.0]$.  Each augmentation is applied with a probability of $0.5$. Figure \ref{fig:data_augmentation} illustrates an example of such augmentation and demonstrates the model's robustness to these augmentations.
 
 \subsection{Implementation Details} The code is implemented using PyTorch \citep{paszkePyTorchImperativeStyle2019}. We used the AdamW optimizer\citep{loshchilovDecoupledWeightDecay2019} with a weight decay of 0.1, batch size of 4, learning rate of $1e-4$ for the prompt predictor network, and $1e-5$ for the mask decoder parameters.  
 $\lambda_1, \lambda_2, \lambda_3$ in Eq. (\ref{eqn:loss_fn}) are set to 10, 1, 1, respectively. The focal loss has parameters $\gamma = 3$ and $\alpha = 0.7$ in Eq. (\ref{eqn:focal_loss}). To maintain the aspect ratio, we resized the longer side of each image to 1024 pixels and padded the shorter side with zeros.

 \subsection{Evaluation Metrics} Models' performance was evaluated using Intersection over Union (IoU) and the Dice similarity coefficient on a held-out test set.  \textbf{Table} \ref{tab:num_train_test} shows the number of test images for each dataset used in this study.
 
 \begin{table}[!tbh]
    \centering
    \resizebox{0.9\textwidth}{!}{%
    \begin{tabular}{l ccc} 
        \toprule
        Model & Hip & Wrist & Shoulder\\
        \midrule
        MobileSAM \citep{zhangFasterSegmentAnything2023a} & 68.2/54.1 & 50.8/35.9  & \textcolor{red}{18.1/10.2} \\
        SAM ViTB & 76.7/62.9  & 59.3/43.7 & \textcolor{red}{22.2/12.9} \\
        MedSAM \citep{maSegmentAnythingMedical2023} & 14.5/8.3 &  20.5/11.8 & \textcolor{red}{13.1/7.2} \\
        SAM ViTH & 75.4/62.5 & 54.7/39.8 & \textcolor{red}{18.5/10.6} \\
        \midrule
        \rowcolor{gray!25}
        SAM2 (Hiera-Tiny)   & 75.1/61.2 & 49.9/35.3 & \textcolor{red}{15.3/8.8} \\
        \rowcolor{gray!25}
        SAM2 (Hiera-Small)  & 74.7/60.9 & 45.6/31.9 & \textcolor{red}{13.0/7.6} \\
        \rowcolor{gray!25}
        SAM2 (Hiera-Large)  & 75.2/61.0 & 59.5/44.3 &  \textcolor{red}{25.2/15.1} \\
        \rowcolor{gray!25}
        SAM2 (Hiera-Base+)  & 78.2/65.2  & 63.5/47.9 &  \textcolor{red}{25.3/15.1} \\
        \bottomrule
    \end{tabular}
    }
    \caption{\textbf{Zero-shot} Dice/IoU scores of SAM variants on three US datasets. }
    \label{tab:zero_shot_performance}
\end{table}

\begin{table}[!t]
     \centering
     \resizebox{0.9\textwidth}{!}{%
     \begin{tabular}{l cc >{\columncolor{gray!25}}c >{\columncolor{gray!25}}c} 
         \toprule
         Model  & \multicolumn{2}{c}{Bounding Box} & \multicolumn{2}{>{\columncolor{gray!25}}c}{Learned Prompts} \\ 
         \cmidrule(lr){2-3} \cmidrule(lr){4-5} 
          & Hip & Wrist & Hip & Wrist \\ 
         \midrule
         SAM2 (Hiera-Tiny)   & 75.1/61.2 & 49.9/35.3 & 81.2/69.0 & 82.7/71.1 \\
         SAM2 (Hiera-Small)  & 74.7/60.9 & 45.6/31.9 & 80.5/68.1 & 83.1/71.6 \\
         SAM2 (Hiera-Large)  & 75.2/61.0 & 59.5/44.3 & 81.6/70.0 & 83.1/71.7  \\
         SAM2 (Hiera-Base+)  & 78.2/65.2  & 63.5/47.9 & 80.4/68.0 & 83.1/71.7 \\
         \bottomrule
     \end{tabular}
     }
     \caption{Comparison of bounding box and learned prompts on hip and wrist datasets using a \textbf{frozen SAM2}. The results demonstrate that learned prompts perform better than bounding box prompts without fine-tuning any SAM2 parameters. \textit{Our results are highlighted in gray. Scores are presented as Dice/IoU}.}
     \label{tab:frozen_performance}
 \end{table}

\begin{table}[!t]
    \centering
    \resizebox{\textwidth}{!}{%
    \begin{tabular}{l c c c c c c}
        \toprule
        & \multicolumn{2}{c}{Hip} & \multicolumn{2}{c}{Wrist} & \multicolumn{2}{c}{Shoulder (bursa)} \\
        \cmidrule(lr){2-3} \cmidrule(lr){4-5} \cmidrule(lr){6-7}
        Model & LP & BB+LP & LP & BB+LP & LP & BB+LP \\ 
        \midrule
        MobileSAM & 78.4/65.7 & 81.4/69.2 & 79.8/67.3 & 81.4/69.3 & 73.7/60.7 & 75.0/62.0 \\
        SAM-ViTB  & 81.9/70.3 & 84.6/73.8 & 82.6/71.0 & 83.9/72.8 & 75.4/62.9 & 76.0/63.5 \\
        \midrule
        \rowcolor{gray!25}
        SAM2Rad (Hiera-Tiny) & 82.4/70.9 & 84.0/73.0 & 83.6/72.4 & 84.6/73.8 & 76.3/63.9 & 77.6/65.3 \\
        \rowcolor{gray!25}
        SAM2Rad (Hiera-Small) & 82.6/71.1 & 84.7/73.9 & 84.5/73.7 & 85.5/75.1 & 74.8/61.9 & 75.6/62.7 \\
        \rowcolor{gray!25}
        SAM2Rad (Hiera-Large) & 81.6/69.9 & 84.3/73.5 & 76.5/64.1 & 77.2/64.5 & 76.5/64.1 & 77.2/64.7 \\
        \rowcolor{gray!25}
        SAM2Rad (Hiera-Base+) & 82.9/71.7 & 85.5/75.2 & 83.7/72.6 & 85.1/75.2 & 75.0/62.1 & 75.6/62.7 \\
        \bottomrule
    \end{tabular}
    }
    \caption{\textit{Sam2Rad Dice/IoU results}. \textbf{LP} - Learned Prompts, \textbf{BB + LP} - Bounding Box plus Learned Prompts.}
    \label{tab:main_results}
\end{table}

 \begin{figure}[h]
     \centering
     \begin{tikzpicture}
         \begin{axis}[
             ybar,
             bar width=15pt,
             width=0.6\textwidth,
             height=8cm,
             symbolic x coords={Sam2-Tiny, SAM2-Small, SAM2-Large, SAM2-Base+},
             xtick=data,
             ymin=0, ymax=1, 
             nodes near coords,
             legend style={at={(0.5,-0.4)}, anchor=north,legend columns=-1},
             enlarge x limits={abs=0.5cm},
             xticklabel style={rotate=45, anchor=east} 
         ]
         
         \addplot coordinates {(Sam2-Tiny,0.78) (SAM2-Small,0.78) (SAM2-Large,0.75) (SAM2-Base+,0.74)};

         \addplot coordinates {(Sam2-Tiny,0.65) (SAM2-Small,0.65) (SAM2-Large,0.63) (SAM2-Base+,0.61)};
         
         \legend{Dice, IoU}
         
         \end{axis}
     \end{tikzpicture}
     \caption{\textit{Prompting SAM2 with few-shot images}. The figure shows the performance of various SAM variants trained on just 10 images and evaluated on the entire test dataset. The proposed prompting strategy demonstrates data efficiency across all SAM variants. }
     \label{fig:few_shot_results}
 \end{figure}
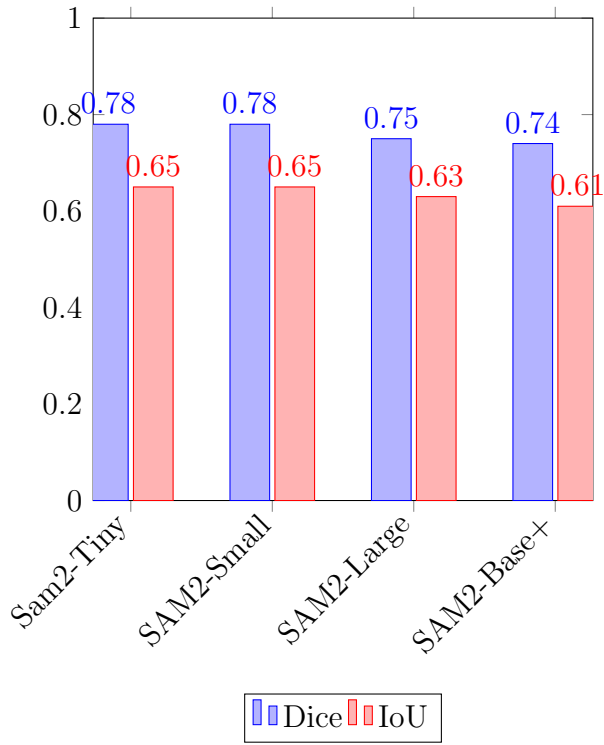

 \subsection{Does prompt learning work any better than bounding box prompts?}
 To evaluate the effectiveness of learned prompts versus bounding box prompts, we conducted an experiment using frozen SAM2. The results, presented in Table \ref{tab:frozen_performance}, demonstrate that learned prompts consistently outperform bounding box prompts across all SAM2 variants. For instance, on the wrist dataset, SAM2 (Hiera-Base+) achieved a Dice score of 83.5 with learned prompts, compared to 63.5 with bounding box prompts.  These findings suggest that the prompt learning strategy not only addresses SAM's need for manual prompts but also introduces additional information not present in manual bounding box prompts. The full results, including LoRA \citep{huLoRALowRankAdaptation2021} fine-tuning the mask decoder, are shown in Table \ref{tab:main_results}.

\subsection{Zero-shot segmentation}
First, we evaluate the zero-shot generalizability of frozen SAM, SAM2, and MedSAM \citep{maSegmentAnythingMedical2023} (fine-tuned on several medical imaging modalities) on three US datasets. We prompt SAM with the ground truth bounding box. The results in Table \ref{tab:zero_shot_performance} show that SAM2 outperforms SAM on all datasets, with the most significant improvement observed on the wrist dataset—although it produces unpredictable results in edge cases. These findings demonstrate that SAM2 aligns better with US images, even without fine-tuning. However, on more challenging datasets such as the shoulder dataset, both SAM and SAM2 perform poorly, indicating the need for fine-tuning. Surprisingly, MedSAM, despite being fine-tuned on medical images, underperforms compared to SAM and SAM2 on US images, even though the latter two were trained solely on natural images.

 \subsection{Few-shot segmentation}
 To further validate the effectiveness of our approach, we trained the model with only 10 images and evaluated it on the corresponding test dataset. The results in Figure \ref{fig:few_shot_results} show that the proposed prompting strategy works well across all SAM variants, with SAM2 having superior performance. The results demonstrate that our approach is effective in the low data regime, making it suitable for medical imaging applications where expert-labeled data is scarce.

\subsection{Robustness to heavy data augmentation}
 Figure \ref{fig:data_augmentation} demonstrates the models' robustness to heavy data augmentation on the test set. This is crucial for generalization to unseen data, especially in overparameterized models like many foundation models. The results suggest that the model is learning the underlying structure of the data rather than simply memorizing the training set.

 \subsection{Robustness to noisy ground truth masks}
 \begin{figure}
    \includegraphics[width=\textwidth]{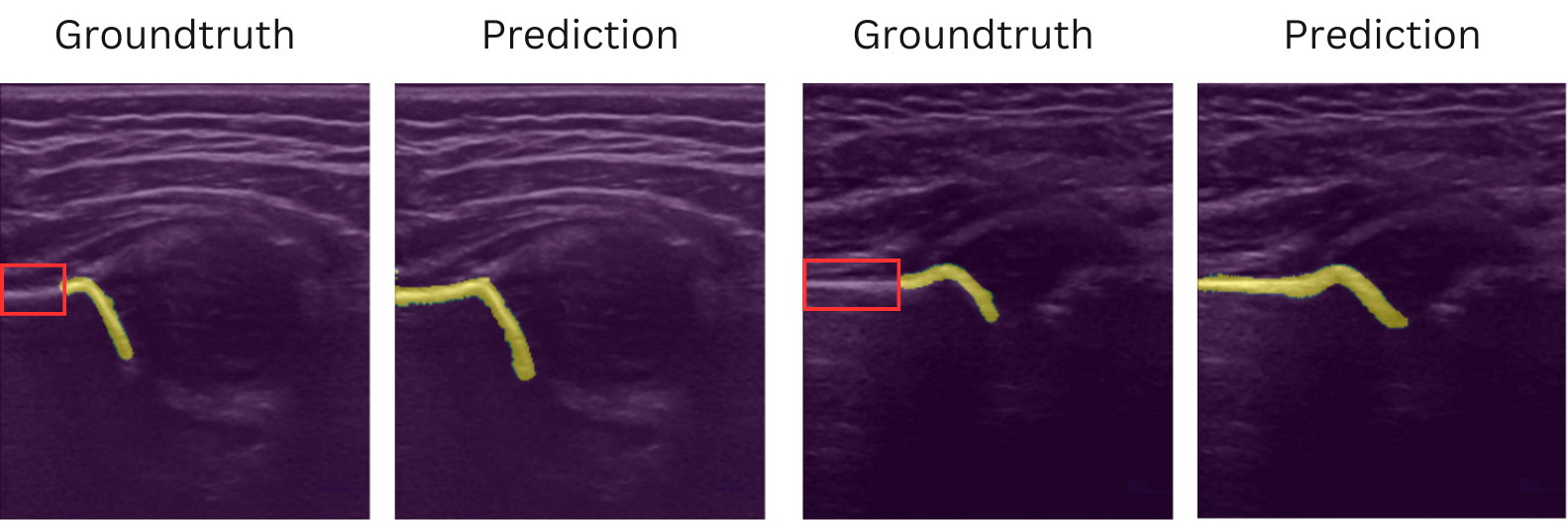}
    \caption{\textit{Sam2-Tiny with PPN predictions on test images}. The model provides complete bone segmentation, in contrast to the partial masks present in the ground truth labels.}
    \label{fig:label_noise}
\end{figure}

Our labels contained various inconsistencies, for instance, partial masks. Despite these imperfections, the model effectively averaged out the noise, producing accurate segmentations on test images. Figure \ref{fig:label_noise} provides visual examples of this robustness.

\section{Conclusion}
In this paper, we presented Sam2Rad, a novel prompt learning approach that significantly enhances the performance of SAM/SAM2 and its variants for ultrasound image segmentation. Our method addresses key limitations of promptable models by eliminating the need for manual prompts and improving segmentation accuracy across various datasets. Sam2Rad outperformed all SAM and SAM2 variants in zero-shot generalization, particularly for challenging cases like shoulder ultrasound. The model demonstrated remarkable improvements, with Dice score increases ranging from 2-7\% for hip and wrist datasets, and up to 33\%  (from 49\% to 82\%) for shoulder data. Notably, Sam2Rad can be trained with as few as 10 labeled images. By leveraging SAM's image encoder features, our network seamlessly integrates with SAM's architecture to produce high-quality prompts. Our model offers three modes: autonomous, semi-autonomous with human-in-the-loop, and fully manual. The seamless integration with SAM's architecture and compatibility with any SAM variant make Sam2Rad a versatile and powerful tool for automatic segmentation in medical imaging.

\section*{Acknowledgement}
 Dr. Jacob L. Jaremko is a Canada CIFAR AI Chair, and his academic time is made available by Medical Imaging Consultants (MIC), Edmonton, Canada. We acknowledge the support of TD Ready Grant, IC-IMPACTS, One Child Every Child, Arthritis Society, and Alberta Innovates AICE Concepts for financial support, the Alberta Emergency Strategic Clinical Network for clinical scanning, and Compute Canada in providing us with computational resources including high-power GPU that were used for training and testing our deep learning models.

\bibliographystyle{elsarticle-num} 
\bibliography{references}

\end{document}